\documentclass{egpubl}
\usepackage{pg2022}

\SpecialIssuePaper         

\usepackage[T1]{fontenc}
\usepackage{dfadobe}  

\usepackage{cite}  
\BibtexOrBiblatex
\electronicVersion
\PrintedOrElectronic
\ifpdf \usepackage[pdftex]{graphicx} \pdfcompresslevel=9
\else \usepackage[dvips]{graphicx} \fi

\usepackage{egweblnk} 
\usepackage{threeparttable}
\usepackage{booktabs}
\usepackage{multirow}
\usepackage{amsmath,amssymb}
\usepackage{hyperref}

\title[TogetherNet: Bridging Image Restoration and Object Detection Together via Dynamic Enhancement Learning]%
      {TogetherNet: Bridging Image Restoration and Object Detection Together via Dynamic Enhancement Learning}

\author[Y. Wang et al.]
{\parbox{\textwidth}{\centering Yongzhen Wang$^{1}$\orcid{0000-0001-6020-3211}, 
         Xuefeng Yan$^{1}$\thanks{Corresponding author (yxf@nuaa.edu.cn).} \orcid{0000-0001-6030-0855},
         Kaiwen Zhang$^{1}$\orcid{0000-0002-6612-5063},
         Lina Gong$^{1}$\orcid{0000-0002-5272-6706},
         Haoran Xie$^{2}$\orcid{0000-0003-0965-3617}, 
         Fu Lee Wang$^{3}$\orcid{0000-0002-3976-0053},
         Mingqiang Wei$^{1,4}$\orcid{0000-0003-0429-490X}
        }
        \\
{\parbox{\textwidth}{\centering $^1$School of Computer Science and Technology, Nanjing University of Aeronautics and Astronautics, Nanjing, China\\
         $^2$Department of Computing and Decision Sciences, Lingnan University, Hong Kong, China\\
         $^3$School of Science and Technology, Hong Kong Metropolitan University, Hong Kong, China\\
         $^4$ Shenzhen Research Institute, Nanjing University of Aeronautics and Astronautics, Shenzhen, China\\
      }
}
}

\begin{document}


\maketitle
\begin{abstract}
 Adverse weather conditions such as haze, rain, and snow often impair the quality of captured images, causing detection networks trained on normal images to generalize poorly in these scenarios. In this paper, we raise an intriguing question – if the combination of image restoration and object detection, can boost the performance of cutting-edge detectors in adverse weather conditions. To answer it, we propose an effective yet unified detection paradigm that bridges these two subtasks together via dynamic enhancement learning to discern objects in adverse weather conditions, called TogetherNet. Different from existing efforts that intuitively apply image dehazing/deraining as a pre-processing step, TogetherNet considers a multi-task joint learning problem. Following the joint learning scheme, clean features produced by the restoration network can be shared to learn better object detection in the detection network, thus helping TogetherNet enhance the detection capacity in adverse weather conditions. Besides the joint learning architecture, we design a new Dynamic Transformer Feature Enhancement module to improve the feature extraction and representation capabilities of TogetherNet. Extensive experiments on both synthetic and real-world datasets demonstrate that our TogetherNet outperforms the state-of-the-art detection approaches by a large margin both quantitatively and qualitatively. Source code is available at \href{https://github.com/yz-wang/TogetherNet}{https://github.com/yz-wang/TogetherNet}.
   


\end{abstract}

\section{Introduction}

Object detection has been widely used in various practical real-world applications \cite{chen2015deepdriving,he2021end,bozcan2021context,wang2022detecting}. Despite the success of learning-based detectors on normal images, they usually fail to detect objects in images with adverse weather conditions, especially in hazy images \cite{chen2018domain,huang2020dsnet}. This can be attributed to the noticeable degradation in image visibility and contrast caused by variant weather, which in turn drops the performance of object detectors. How to improve the accuracy of cutting-edging detectors in adverse weather conditions has attracted a great deal of attention \cite{sindagi2020prior,hnewa2021multiscale,liu2022image,sun2021multi}.

\begin{figure}[!h] \centering
	\includegraphics[width=1.0\linewidth]{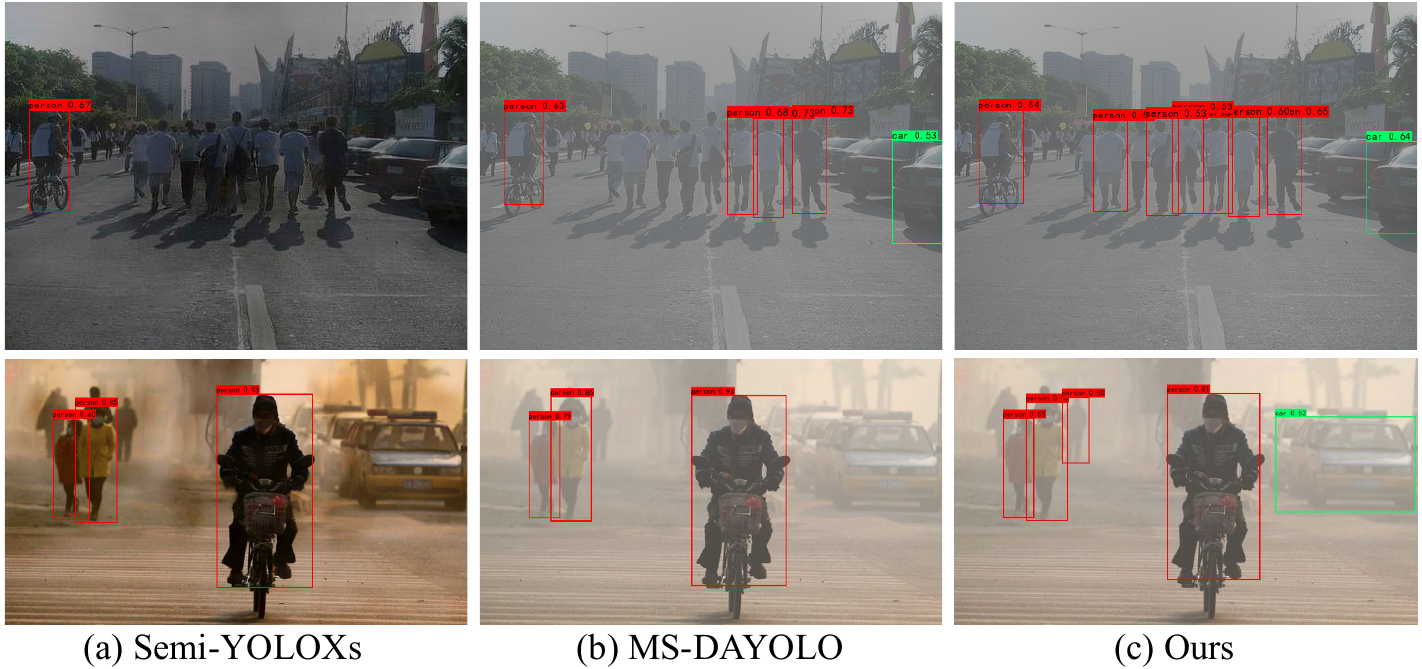}
	\caption{
	Detection results by different methods on two typical examples of adverse weather conditions. From (a) to (c): the detection results by (a) Semi-YOLOXs \cite{li2019semi} ("dehaze+detect"), (b) MS-DAYOLO \cite{hnewa2021multiscale}, and (c) our TogetherNet. 
	}
	\label{fig111}
\end{figure}

To tackle this challenging problem, an intuitive solution is to mitigate the effects of weather conditions by pre-processing the images using the restoration techniques such as image dehazing \cite{liu2019griddehazenet,li2019semi,dong2020multi,ren2020scga,li2021dehazeflow}. Most of them can enhance the overall visibility of these degraded images. But, can the restored images serve the downstream object detection task effectively? The answer may be not positive. That is because the restored images lose some important details which are beneficial to object detection \cite{li2018benchmarking}. \textit{We consider the good way to serve object detection is to make object detection itself involve image restoration.} However, if simply combining a dehazing network with a detection network in a cascaded manner will increase the computational overhead and slow the inference time, which is undesirable in resource-constrained applications. \textit{We further consider the good way to serve object detection in adverse conditions is to bridge image restoration and object detection together in a unified yet joint learning paradigm.} 

In this paper, we respond to the intriguing learning-related question: combining a low-level image restoration task with a high-level object detection task to develop a multi-task joint learning paradigm will improve the performance of cutting-edge detection models in adverse weather conditions. Accordingly, we propose a novel unified paradigm that bridges these two subtasks together via dynamic enhancement learning for discerning objects in adverse weather conditions, termed a TogetherNet. Specifically, TogetherNet employs a cutting-edge object detector (i.e., YOLOX \cite{ge2021yolox}) as the detection module, and exploits a feature restoration module to share the feature extraction module (backbone) with the detection network for image restoration. We train TogetherNet in an end-to-end fashion to simultaneously learn about image restoration and object detection. In this way, the latent information hidden in degraded images can be restored to benefit the detection task. In turn, the training of the detection task helps the backbone network to extract deeper structural and detailed features, thus facilitating the image restoration task. Moreover, considering that the performance of object detectors under adverse weather is usually limited, a Dynamic Transformer Feature Enhancement module (DTFE) is proposed to further enhance the feature extraction and representation capabilities of the model, thus improving its detection accuracy in such scenarios.

Recently, some approaches \cite{saito2019strong,liu2022image,hnewa2021multiscale,rezaeianaran2021seeking} cast object detection in adverse weather conditions as a task of learning models from a source domain (clean images) to a target domain (under adverse weather), i.e., unsupervised domain adaptation. These methods consider that compared with the clean images (source domain) used to train the detectors, the images (target domain) captured in adverse weather suffer from an obvious domain shift problem \cite{gopalan2011domain,chen2018domain}. They mostly employ domain adaptation strategies such as adversarial training to alien the target features with the source features. Despite the promising results achieved by domain adaptation under adverse weather, they usually ignore the latent information hidden in the degraded images which can also provide additional beneficial information for the detection task. As exhibited in Figure \ref{fig111}, compared with the $``dehaze+detect"$, and domain adaptive-based detection models, the proposed TogetherNet can detect more objects with higher confidence, which demonstrates that our model outperforms the other algorithms for detecting objects in adverse weather. Note that Semi-YOLOXs is a typical $``dehaze+detect"$ method that first restores the image and then detects the object, so the results in Figure \ref{fig111} look different from other methods.

Extensive experiments on both synthetic benchmark (VOC-FOG-test) and real-world datasets (Foggy Driving Dataset \cite{sakaridis2018semantic} and RTTS \cite{li2018benchmarking}) demonstrate that our TogetherNet is far superior to the state-of-the-art object detection approaches. In summary, the main contributions are threefold:
\begin{itemize}
	\item An effective yet unified detection paradigm is proposed for discerning objects in adverse weather conditions, which leverages a joint learning framework to perform image restoration and object detection tasks simultaneously, called, TogetherNet.
	\item We propose a Dynamic Transformer Feature Enhancement module (DTFE) to enhance the feature extraction and representation capabilities of TogetherNet.  
	\item We compare TogetherNet with various representative state-of-the-art object detection approaches via extensive experiments, including $``dehaze+detect"$, domain adaptive-based, multi-task-based, and image adaptive-based detection models. Consistently and substantially, TogetherNet performs favorably against them.
\end{itemize}

\section{Related Work}
In this section, we briefly summarize state-of-the-art object detectors that have produced encouraging results in general scenarios and under adverse weather conditions.

\subsection{Object Detection}
As a long-standing and fundamental task in computer vision, object detection has attracted extensive research attention in academia and industry \cite{zhao2019object,liu2020deep}. Recently, with the rapid development of convolutional neural networks (CNNs), learning-based detectors have dominated the modern object detection field for years. 

Current object detection methods can be broadly categorized into two major groups, namely region proposal-based and regression-based. For region proposal-based approaches, they typically first employ methods such as selective search \cite{uijlings2013selective} to produce the candidate proposals, and then refine them for subsequent object detection. R-CNN \cite{girshick2014rich} is the most representative region proposal-based detector, which adopts a CNN to extract features for the produced proposals, and then applies a support vector machine to perform classification. Inspired by the success of R-CNN in object detection, numerous variants based on this framework have sprung up, including Fast R-CNN \cite{girshick2015fast}, Faster R-CNN \cite{ren2015faster}, Libra R-CNN \cite{pang2019libra} and Dynamic R-CNN \cite{zhang2020dynamic}. Despite achieving encouraging detection accuracy, region proposal-based approaches are not satisfactory in terms of inference speed, which are undesirable in real-time applications. Therefore, to achieve a better speed-performance trade-off, various regression-based methods are developed for real-time detection. Representative approaches including YOLO series \cite{redmon2016you, redmon2017yolo9000,redmon2018yolov3,bochkovskiy2020yolov4,yolov5,ge2021yolox}, SSD \cite{liu2016ssd}, RetinaNet \cite{lin2017focal}, CenterNet \cite{zhou2019objects}, etc. In a nutshell, regression-based detectors are generally faster, but their detection performance is slightly weaker than region proposal-based detectors.

\begin{figure*}[!h] \centering
	\includegraphics[width=1.0\linewidth]{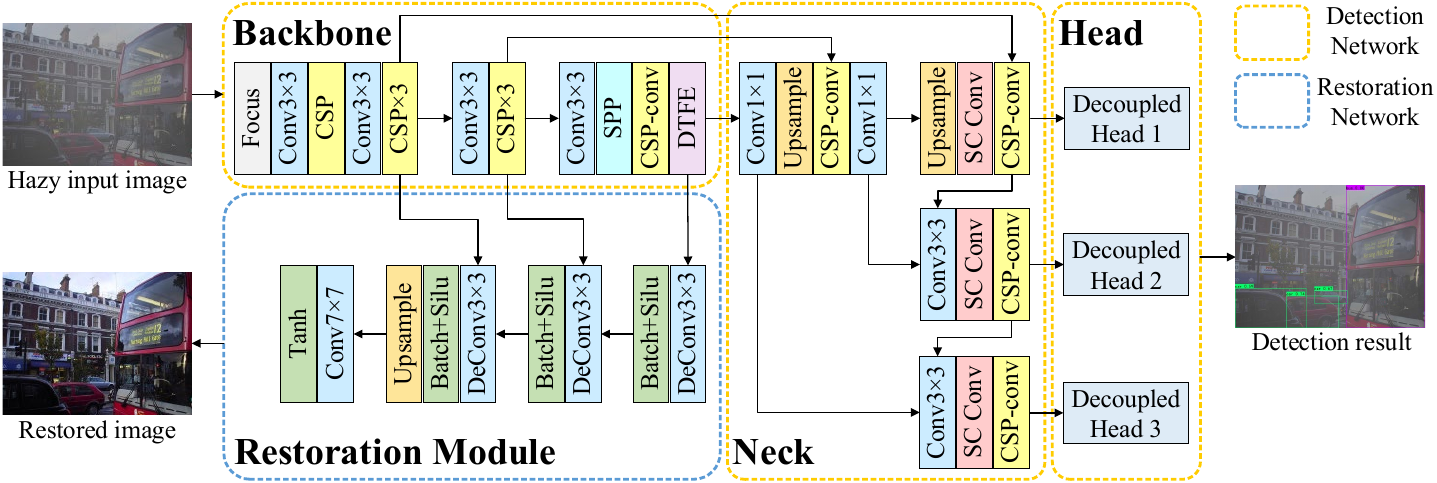}
	\caption{
	The architecture of TogetherNet. It consists of both the object detection and image restoration networks. Note that the restoration module is only activated during the training phase. CSP/CSP-conv refers to Cross Stage Partial Network \cite{wang2020cspnet} with/without residual networks, DTFE refers to Dynamic Transformer Feature Enhancement module, and SC Conv refers to self-calibrated convolutions \cite{liu2020improving}. }
	\label{fig:fig1}
\end{figure*}

\subsection{Object Detection in Adverse Weather}
Compared with general object detection, few research efforts have been explored on object detection in adverse weather conditions. Early methods mainly focused on pre-processing the degraded images by existing restoration algorithms such as image dehazing \cite{he2010single,qin2020ffa,wang2022cycle,sun2022sadnet} or image deraining \cite{liang2019rain,ren2020not,DBLP:conf/cvpr/DengWWFL0WW20}, and then sending the processed images to the subsequent detection network for object detection. Although employing image restoration approaches as a pre-processing step can improve the overall quality of degraded images, these images may not definitely benefit the detection performance. A few prior-based efforts \cite{li2017aod,huang2020dsnet} have attempted to jointly perform image restoration and object detection to mitigate the effects of adverse weather-specific information. Sindagi et al. \cite{sindagi2020prior} develop a prior-based unsupervised domain adaptive framework for detecting objects in hazy and rainy conditions. Liu et al. \cite{liu2022image} propose an image-adaptive detection network for object detection in adverse weather conditions, which combines image restoration and object detection into a unified framework and achieves very promising results. Recently, several methods \cite{shan2019pixel,zhu2019adapting,roychowdhury2019automatic,zhang2021domain} have begun to exploit domain adaption to overcome this problem. Zhang et al. \cite{zhang2021domain} treat object detection in adverse weather as a domain shift problem and propose a domain adaptive YOLO to improve cross-domain performance for one-stage detectors.

\section{TogetherNet}
To boost the detection capacity of object detectors in adverse weather conditions, we come up with such a solution - is it possible to develop a unified detection paradigm that incorporates such detection task as a joint learning framework, while encouraging image restoration and detection tasks to benefit each other? If the answer is positive, our method can effectively address the detection of adverse weather scenarios. It is the focus of this work. 

In this section, we first introduce the overview of the proposed unified detection paradigm, called TogetherNet, to demonstrate how we address the detection problem under adverse weather conditions. Next, the proposed restoration network is described in detail. After that, we elaborate on the proposed Dynamic Transformer Feature Enhancement module (DTFE) to explore the potential of deformable convolutions and self-attention mechanisms in feature extraction and representation. Finally, we describe the self-calibrated convolutions and Focal loss for optimizing our network to further improve the detection performance in adverse weather.

\subsection{Overview of TogetherNet}
The overall architecture of the proposed TogetherNet is depicted in Figure \ref{fig:fig1}. Different from existing detection efforts, we consider overcoming the detection task from the following three perspectives. First, we employ an image restoration module to mitigate the influence of weather-specific information on the detection task. Second, a multi-task joint learning paradigm is developed to encourage low-level image restoration and high-level object detection tasks to collaborate and promote each other. Finally, a feature enhancement module is exploited to improve the feature extraction and representation capabilities of the model, such that more latent features can be revealed from degraded images to benefit image restoration and detection tasks.

YOLO series detectors \cite{redmon2016you, redmon2017yolo9000,redmon2018yolov3,bochkovskiy2020yolov4,yolov5,ge2021yolox} are the most representative regression-based detection models, which have been successfully applied in numerous scenarios. Recently, YOLOX \cite{ge2021yolox} has been released as the latest version of the YOLO series detectors. Despite the promising results achieved by YOLOX in various benchmark datasets (e.g., MSCOCO \cite{lin2014microsoft}, PASCAL-VOC \cite{everingham2010pascal}), there are still many challenging yet unsolved problems. First, the YOLOX family detectors are originally designed for object detection in general yet easy scenes, without considering how to cope with the object detection in adverse weather conditions. Second, similar to most existing detectors, the YOLOX family detectors are susceptible to the weather-specific information in the detection task under adverse weather, resulting in a significant drop in detection accuracy. Third, YOLOXs (the smallest version of the YOLOX family) is very lightweight and efficient, which is promising for resource-constrained mobile devices. But its detection performance is thus largely dropped for the adverse weather scenes. 

To this end, we start from these three aspects and propose a novel unified detection paradigm for discerning objects in adverse weather conditions, called TogetherNet. To attain this objective, the proposed TogetherNet adopts one of the best object detectors, i.e., YOLOXs as our detection network to perform detection task. Our TogetherNet has the potential to benefit from a more complex version of the YOLOX family (e.g., YOLOXm and YOLOXl) to further improve its detection performance. However, we choose the smallest version of YOLOX because a lightweight model is more desirable in resource-limited/real-time applications.

As depicted in Figure \ref{fig:fig1}, the proposed TogetherNet consists of two main modules, i.e., the detection network and the restoration network. Given a hazy input image, we first employ the focus operation in the backbone module to separate the image into different granularities and regroup them together to enhance the image features. Then, several cross stage partial modules (CSP) \cite{wang2020cspnet} and a Dynamic Transformer Feature Enhancement module (DTFE) are employed to extract the complex and latent features from the restructured feature map. DTFE is a novel feature enhancement module developed to expand the receptive field with adaptive shape and enhance the model’s feature representation capability for better detection and image restoration. After that, the extracted features are transmitted to both the restoration module and the neck module to perform different tasks. In this way, TogetherNet can benefit from the joint learning framework, where the clean features produced by the restoration module can be shared to learn better object detection in the detection network. Finally, the detection head module will produce the final class probability scores, bounding boxes, and confidence scores.

Moreover, to further improve the detection capacity of TogetherNet and well address the challenge of detecting objects in adverse weather, we introduce a multi-scale feature enhancement module, namely, self-calibrated convolutions and the well-known Focal loss into our model. Both of them have been widely used in object detection networks and proved to be effective in improving detection accuracy, which will be described in the following sections. We emphasize that, since this work mainly focuses on object detection in adverse weather, and introducing the image restoration module in the testing phase would significantly slow down TogetherNet's inference speed, the restoration module is only activated during the training phase.

\begin{figure}[htb]
	\centering
	\includegraphics[width=1.0\linewidth]{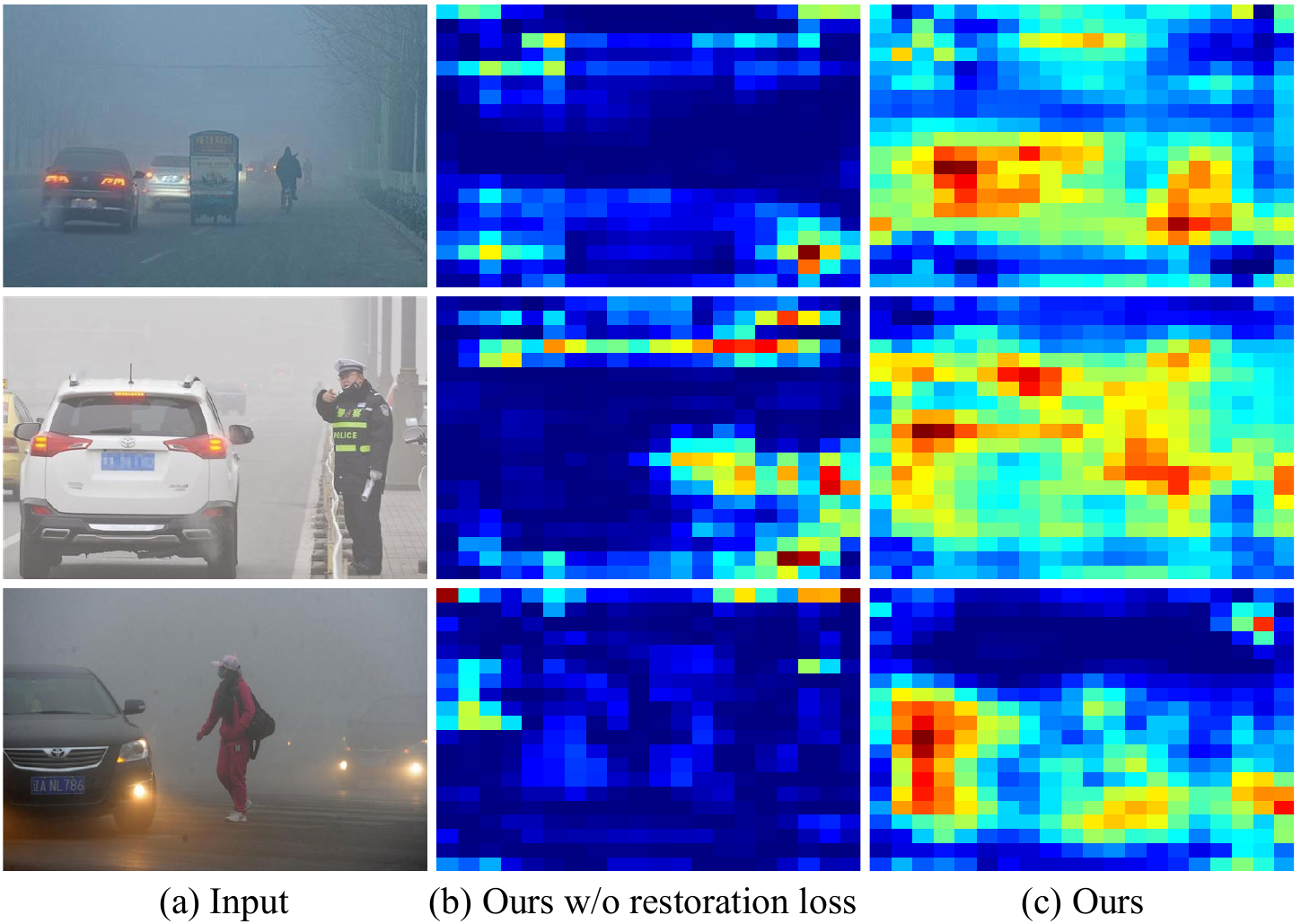}
	\caption{Visualization of feature maps.}
	\label{feature}
\end{figure}

\subsection{Restoration Network}
 In our design, the restoration network is responsible for recovering the clean images and sharing these restored features with the detection network during joint learning to promote the model's detection accuracy in adverse weather conditions. To attain this objective, we employ the backbone network to extract the complex and latent features hidden in the input image for simultaneously learning image restoration and object detection. 
 
 Considering that the features extracted by the backbone network may be degraded by weather-specific information (i.e. haze), resulting in poor detection performance, a decoder-like network is developed as our restoration module to eliminate these effects as much as possible. As demonstrated in Figure \ref{fig:fig1}, to restore the clean image features, three deconvnet, an up-sampling operation, and a $Tanh$ activation function are adopted to produce the final clean images. Moreover, we also introduce the skip connection strategy to facilitate the detection task by revealing multi-scale latent features while avoiding the gradient vanishing problem.
 
 To perform image restoration, the mean square error (MSE) loss is employed to train the restoration network, which can be expressed by:
 \begin{equation}
 L_{re}=\frac{1}{n} \sum_{i=1}^{n}\left(\hat{\mathbf{Y}}_{i}-\mathbf{Y}_{i}\right)^{2},
 \end{equation}
 where $n$ denotes the batch size, $\hat{\mathbf{Y}}_{i}$ refers to the ground truth image, and $\mathbf{Y}_{i}$ refers to the estimated clean image. Actually, using a more complex network architecture or loss function may enhance the dehazing performance of current models, we prefer to adopt a simple CNN-based network and MSE loss to achieve a better parameter-performance trade-off.
 
 To better understand the effectiveness of the proposed restoration network, we visualize the features of the last layer in our backbone module (with/without restoration loss). As depicted in Figure \ref{feature}, features with restoration loss mitigate the influence of weather information on them to some extent, and can still focus on regions containing objects (see the red regions in Figure \ref{feature}), allowing for better performing detection tasks.
 
 Furthermore, we have tried to send the clean images recovered by the restoration network directly to the detection module for object detection, but the detection accuracy appears to be greatly dropped. We argue that the restored clean image weakens some features of the original image, and even creates a new domain shift problem during the image restoration, prohibiting such a strategy from achieving optimal performance. In light of this, we consider employing the restoration network to produce the latent clean features from the backbone network through learning the image restoration task. As a consequence, the improved detection performance of TogetherNet can be achieved by jointly optimizing image restoration and object detection.

\subsection{Dynamic Transformer Feature Enhancement Module}
\begin{figure}[htb]
	\centering
	\includegraphics[width=0.65\linewidth]{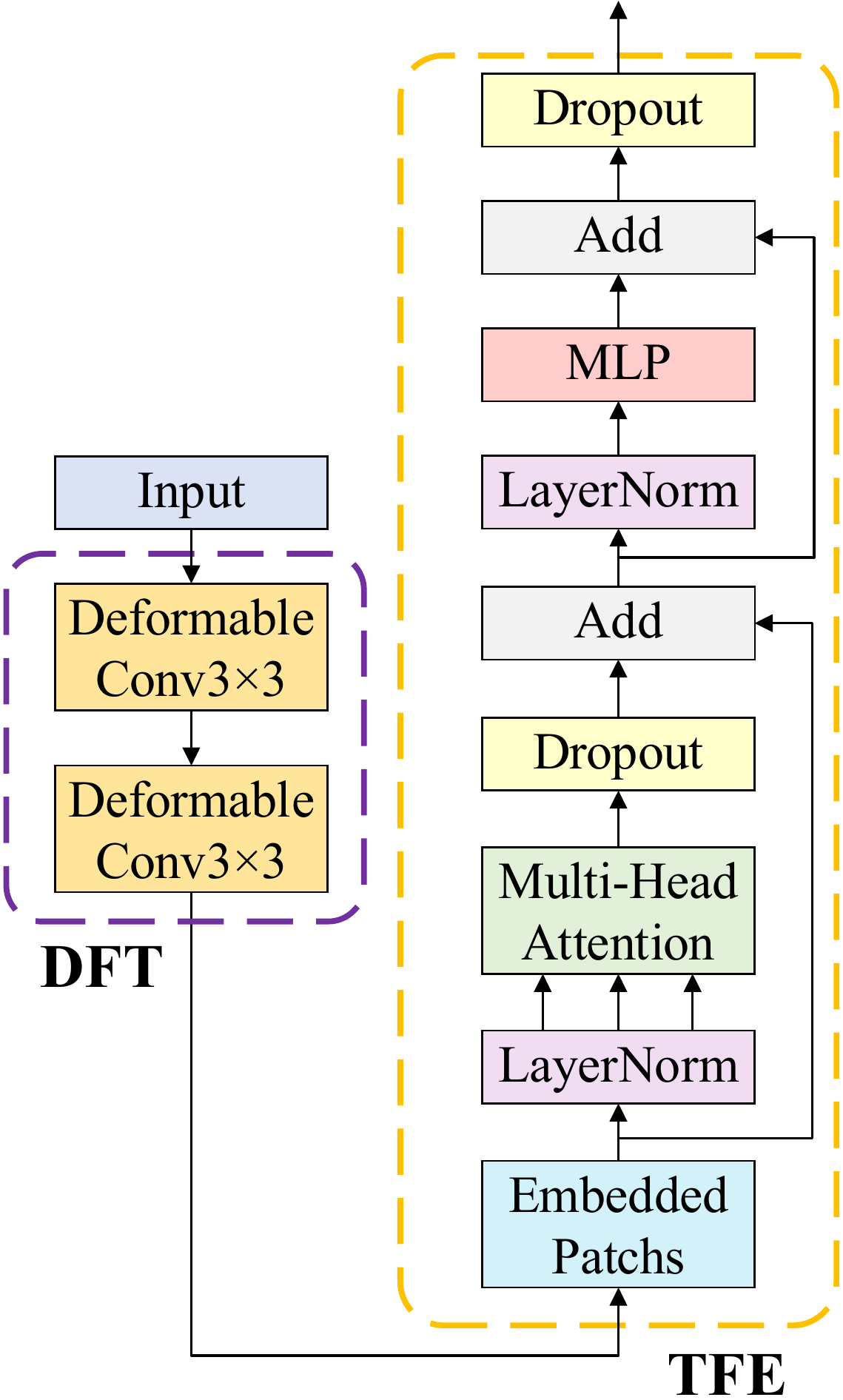}
	\caption{The architecture of the Dynamic Transformer Feature Enhancement module (DTFE). DTFE can help the backbone network improve its feature extraction and representation capabilities for better image restoration and object detection. DFT refers to dynamic feature transformation network, TFE refers to Transformer-based feature enhancement network.}
	\label{dtfe}
\end{figure}

 Fundamentally, the feature extraction and representation capabilities of the network directly determine the performance of the model. Hence, we argue that there are two solutions to reduce the impact of weather-degradations on detection tasks under adverse weather. The first approach is to expand the receptive field of the network to help the model fuse more spatially structured information, so that the objects can be discerned from the area less affected by weather-specific information. Another approach is to enhance the feature extraction capability of the network so that objects can be detected directly from these areas with poor visibility. To this end, we develop a novel Dynamic Transformer Feature Enhancement module (DTFE) to improve the model’s feature extraction and representation capabilities for better image restoration and object detection. The DTFE module mainly consists of two parts, i.e., a dynamic feature transformation network (DFT) and a Transformer-based feature enhancement network (TFE), as depicted in Figure \ref{dtfe}. Specifically, we employ two deformable convolutions \cite{dai2017deformable} to form the dynamic feature transformation network, which can expand the receptive field of the model with adaptive shape and improve its transformation ability. For the feature enhancement network, we adopt the Vision Transformer block \cite{dosovitskiy2020image} to explore the potential of the self-attention mechanisms in improving the model's feature representation capability. 

 Different from conventional CNNs, the kernels in deformable convolutions are dynamic and flexible, which can capture more spatially structured information. Moreover, the work \cite{zhu2020deformable} has demonstrated that dynamic and flexible convolution kernels can effectively improve the feature transformation capabilities of networks, such that the deformable convolutions are employed to enhance the feature for object detection. Therefore, we develop a dynamic feature transformation network based on deformable convolutions to expand the receptive field with adaptive shape and improve the model’s transformation capability (see Figure \ref{dtfe}). In this way, our model can focus on more areas that are less affected by weather-specific information, thus reducing the impact of weather-degradations on detection accuracy. 

 Recently, Vision Transformers (ViTs) have become one of the dominant models in computer vision owing to their ability to learn complex dependencies between input features via self-attention mechanisms. Given this, we consider adopting the ViT module in the design of the backbone network to boost its feature representation ability, thus improving the performance of the subsequent image restoration and object detection tasks. In particular, we introduce a feature enhancement network via the Vit module in the last layer of the backbone network to further enhance the extracted features. It enables the backbone network to build complex and long-range spatial dependencies between the input features, thus improving the detection capacity of our TogetherNet.

\subsection{Self-calibrated Convolutions}
The self-calibrated convolution network is an improved CNN structure proposed by Liu et al. \cite{liu2020improving}, which can build long-range spatial and inter-channel dependencies around each spatial location. Therefore, it can enlarge the receptive field of each convolutional layer and enhance the feature extraction ability of CNNs. In light of this, we consider adopting the self-calibrated convolution network as a multi-scale feature extraction module to cope with the weather-degradation problem in the detection task and improve the detection performance of TogetherNet. 

\begin{figure}[htb]
	\centering
	\includegraphics[width=1.0\linewidth]{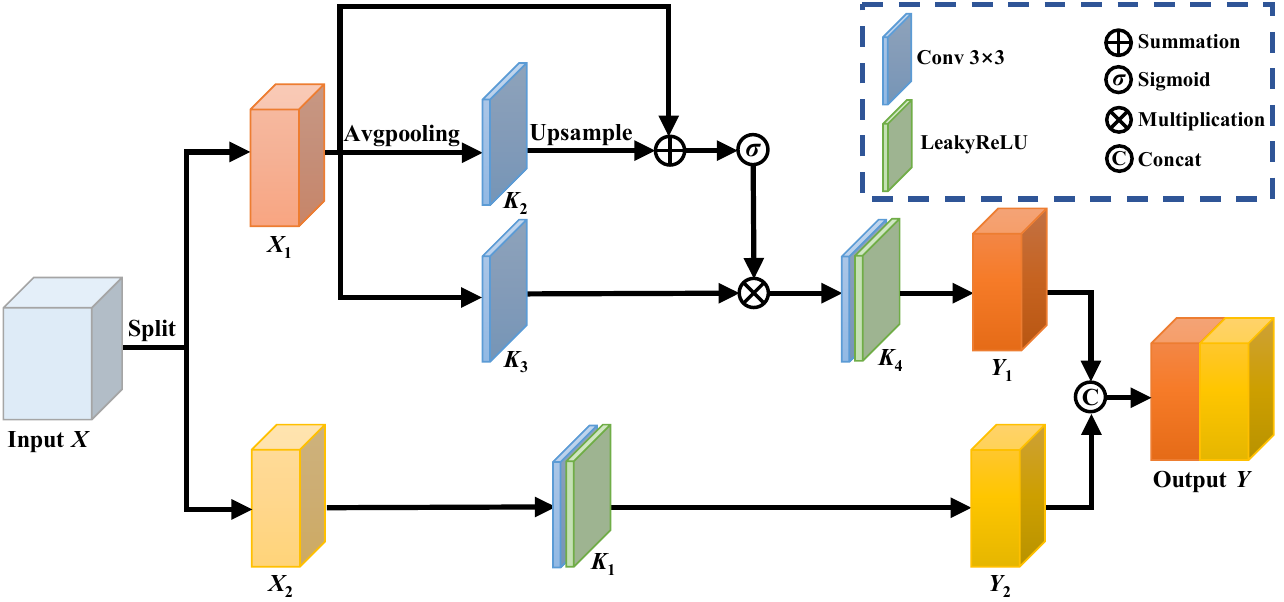}
	\caption{The architecture of the self-calibrated convolutions. It can build long-range spatial and inter-channel dependencies around each spatial location, thus expanding the receptive field of each convolutional layer and enhancing the feature extraction capability of CNNs.}
	\label{fig4}
\end{figure}

The architecture of the self-calibrated convolution network is exhibited in Figure \ref{fig4}. Given an input feature map $X$ with channel $C$, we first split it into two feature maps $X_1$ and $X_2$ with channel $C/2$. Then, we send $X_1$ to the self-calibrated branch for feature transformation and fusion. In this branch, three filters ($K_2$, $K_3$, and $K_4$) are employed to extract and fuse multi-scale features from $X_1$. Next, we employ a filter $K_1$ to transmit and extract features from $X_2$ to obtain the other half of the result $Y_2$. Finally, $Y_1$ and $Y_2$ are concatenated to produce the final output $Y$. In our designation, we introduce self-calibrated convolutions in front of the three decoupled YOLO heads to expand the receptive field of the convolutional layer and extract multi-scale features for better object detection (see Figure \ref{fig:fig1}). In this way, our TogetherNet can well address the challenge of discerning objects in adverse weather conditions.

Moreover, inspired by RetinaNet \cite{lin2017focal} with high effective Focal loss, we introduce the Focal loss in the design of our TogetherNet to address the problem of imbalance between positive and negative samples in object detection tasks. Therefore, the total loss function can be formulated as:
 \begin{equation}
 L_{{Total }}=L_{de}+L_{re},
 \end{equation}
where $L_{de}$ refers to the detection loss, which can be expressed by:
\begin{equation}
L_{de}=\lambda L_{IoU}+L_{Cls}+L_{Focal},
 \end{equation}
 where $\lambda$ is loss weight, and we set $\lambda = 5$ here. $L_{IoU}$ and $L_{Cls}$ refer to regression loss and classification loss, respectively.
 
 In our experiments, we are surprised to find that the image restoration loss $L_{re}$ is very helpful in improving the performance of the detection task, here we set the loss weight of $L_{re}$ to 0.8, and the loss weight of $L_{de}$ to 0.2. For these two loss weights, extensive experiments are performed to ensure their optimum values (see Section \ref{ablation}). Therefore, developing a joint learning paradigm that combines these two tasks is very effective to improve detection capacity in adverse weather conditions.

\section{Experiments}
In this section, comprehensive experiments are performed to evaluate the detection performance of TogetherNet and other detection approaches under adverse weather conditions. To conduct experiments, a dataset for detecting objects in foggy weather is established, called VOC-FOG. For evaluation, Both the synthetic foggy dataset (VOC-FOG-test) and real-world foggy datasets (Foggy Driving dataset \cite{sakaridis2018semantic} and Real-world Task-driven Testing Set (RTTS) \cite{li2018benchmarking}) are employed as the testing set. All the experiments are implemented by PyTorch 1.9 on a system with an Intel(R) Core(TM) i7-9700 CPU, 16 GB RAM, and an NVIDIA GeForce RTX 3090 GPU.

\begin{table}[htbp]
\centering
\footnotesize
	\caption{Details about training and testing dataset. The object classes are bic (bicycle), bus, car, mot (motorcycle), and per (person). FDD is the abbreviation of Foggy Driving Dataset.}
	\label{table1}
	\begin{threeparttable}
	\centering
	\setlength{\tabcolsep}{1.9mm}{
			\begin{tabular}{cccccccc}
				\toprule
				\multirow{1}{*}{Dataset}&
                \multirow{1}{*}{Images} &
                \multirow{1}{*}{Bic} &
                \multirow{1}{*}{Bus} &
                \multirow{1}{*}{Car} &
                \multirow{1}{*}{Mot} &
                \multirow{1}{*}{Per} \cr
				\midrule
				VOC-FOG & 9578 & 836 & 684 & 2453 & 801 & 13519\cr
				VOC-FOG-test & 2129 & 155 & 156 & 857 & 131 & 3527\cr
				FDD & 101 & 17 & 17 & 425 & 9 & 269\cr
				RTTS & 4332 & 534 & 1838 & 18413 & 862 & 7950\cr
				\bottomrule
			\end{tabular}
	}
	\end{threeparttable}
\end{table}
\begin{figure}[!h] \centering
	\includegraphics[width=0.95\linewidth]{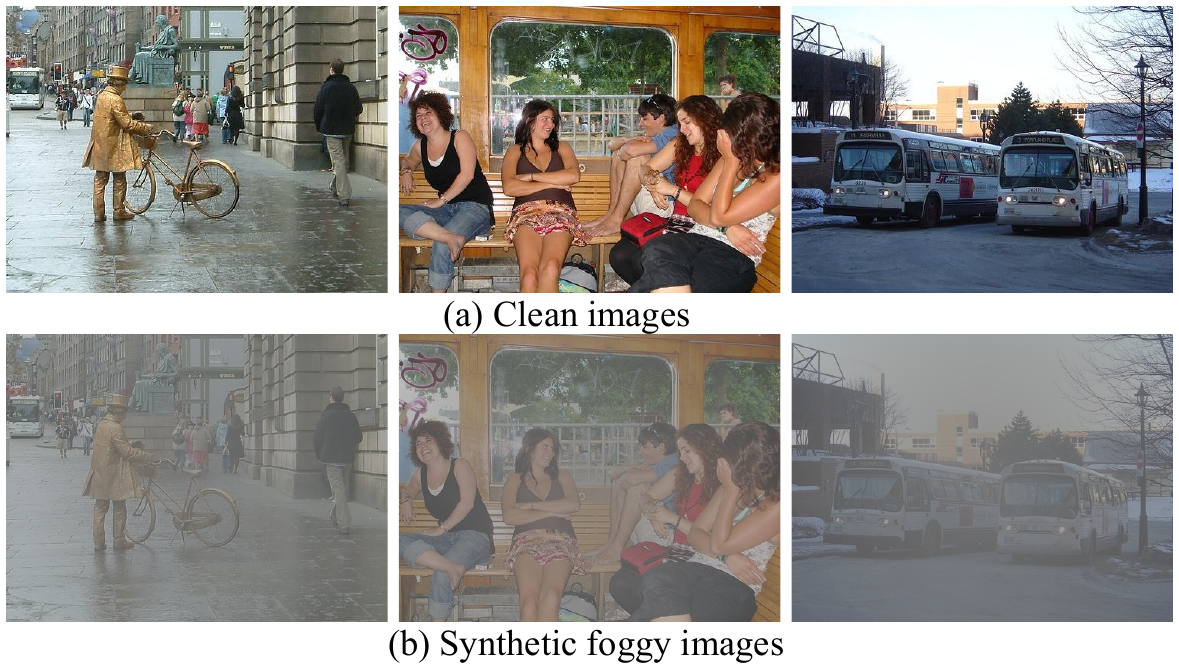}
	\caption{
	Example images in the proposed VOC-FOG dataset. }
	\label{fig13}
\end{figure}

\subsection{Dataset}
Considering there are few publicly available datasets for object detection in adverse weather conditions, to train and evaluate the proposed TogetherNet, we establish a foggy detection dataset based on the classic VOC dataset \cite{everingham2010pascal}, dubbed VOC-FOG. Specifically, we employ the well-known atmospheric scattering model to generate the foggy images $I\left ( x \right )$, which can be obtained by the following formula:
\begin{equation}
  I\left ( x \right )=J\left ( x \right )t\left ( x \right )+A\left ( 1-t\left ( x \right ) \right ),
  \label{eq:3}
\end{equation}
where $J(x)$ denotes the clean image, $A$ refers to the global atmospheric light, and $t(x)$ refers to the medium transmission map, which can be calculated by:
 \begin{equation}
  t\left ( x \right )=e^{-\beta d\left ( x \right )},
  \label{eq:4}
\end{equation}
where $\beta $ denotes the atmosphere scattering parameter, and $d\left ( x \right )$ refers to the scene depth, which can be defined as: 
 \begin{equation}
  d(x) = -0.04\times\rho +\sqrt{max(w,h)},
  \label{eq:8}
\end{equation}
where $\rho$ denotes the Euclidean distance from the current pixel to the central pixel, $w$ and $h$ refer to the numbers of rows and columns of the image. In our experiments, we set the global atmospheric light parameter $A$ to 0.5, while randomly setting the atmospheric scattering parameter $\beta$ between 0.07 and 0.12 to control the fog level. Moreover, considering there are five annotated object classes (i.e., car, bus, motorcycle, bicycle, and person) in the RTTS dataset, to form our training dataset, we select the images containing these five categories to add haze. After processing these clean images on the original VOC dataset, we obtain 9578 foggy images for training (VOC-FOG) and 2129 images for testing (VOC-FOG-test), as depicted in Table \ref{table1} and Figure \ref{fig13}.

\begin{table*}[]
\footnotesize
  \caption{Comparison of TogetherNet with state-of-the-art detection models on the VOC-FOG-test dataset. * denotes that the model is trained with clean images from the VOC-FOG dataset. \textcolor{red}{Red} and \textcolor{blue}{blue} colors are used to indicate the $1^{st}$ and $2^{nd}$ ranks, respectively.}
  \centering
  \begin{tabular}{ccccccccc}
    \toprule
    Method & Publication & Type & Person & Bicycle & Car & Motorbike & Bus & $mAP$ \\
    \midrule
    YOLOXs \cite{ge2021yolox} & arXiv'21 & Baseline & 80.81 & 74.14 & 83.63 & 75.35 & 86.40 & 80.07 \\
    YOLOXs* \cite{ge2021yolox} & arXiv'21 & Baseline & 79.97 & 67.95 & 74.75 & 58.62 & 83.12 & 72.88 \\
    DCP-YOLOXs* \cite{he2010single} & TPAMI’11 & Dehaze & 81.58 & \textcolor{red}{78.80} & 79.75 & 78.51 & 85.64 & 80.86 \\
    AOD-YOLOXs* \cite{li2017aod} & ICCV’17 & Dehaze & 81.26 & 73.56 & 76.98 & 71.18 & 83.08 & 77.21\\
    Semi-YOLOXs* \cite{li2019semi} & TIP’20 & Dehaze & 81.15 & 76.94 & 76.92 & 72.89 & 84.88 & 78.56\\
    FFA-YOLOXs* \cite{qin2020ffa} & AAAI’20 & Dehaze & 78.30 & 70.31 & 69.97 & 68.80 & 80.72 & 73.62\\
    MS-DAYOLO \cite{hnewa2021multiscale} & ICIP’21 & Domain adaptive & \textcolor{blue}{82.52} & 75.62 & \textcolor{red}{86.93} & \textcolor{blue}{81.92} & \textcolor{blue}{90.10} & \textcolor{blue}{83.42}\\
    DS-Net \cite{huang2020dsnet} & TPAMI’21 & Multi-task & 72.44 & 60.47 & 81.27 & 53.85 & 61.43 & 65.89\\
    IA-YOLO \cite{liu2022image} & AAAI’22 & Image adaptive & 70.98 & 61.98 & 70.98 & 57.93 & 61.98 & 64.77\\
    TogetherNet  & ours & Multi-task & \textcolor{red}{87.62} & \textcolor{blue}{78.19} & \textcolor{blue}{85.92} & \textcolor{red}{84.03} & \textcolor{red}{93.75} & \textcolor{red}{85.90}\\
    \bottomrule
  \end{tabular}
  \label{tab2}
\end{table*}

As observed, the fog in the central area appears to be thicker than in the surrounding areas, which can be explained by the principle of synthetic fog. When employing the atmospheric scattering model to generate fog, we first need to select a point as the starting point and then spread the synthetic fog around it. Considering that the center of natural images is generally the position with the largest depth value, it is common to use the center point as the starting point when synthesizing fog. Therefore, the fog in the center is usually thicker.

\textbf{Testing set}. To evaluate the detection performance of TogetherNet and other detection methods in adverse weather conditions, both the synthetic foggy dataset (VOC-FOG-test) and two real-world foggy datasets (Foggy Driving dataset and RTTS) are employed as our testing set.

\begin{itemize}
\item \textbf{VOC-FOG-test} contains 2129 foggy images synthesized from the clean images in the VOC dataset. Different from the above-mentioned VOC-FOG training set, to further verify the generalization ability of TogetherNet, we set atmosphere scattering parameter $\beta$ to a wider range to simulate extreme foggy and misty weather conditions. Specifically, the value of $\beta$ is randomly set between 0.05 and 0.14 to adjust for different fog levels.

\item \textbf{Foggy Driving Dataset} \cite{sakaridis2018semantic} is a real-world foggy dataset that is used for object detection and semantic segmentation. It involves 466 vehicle instances (i.e., car, bus, train, truck, bicycle, and motorcycle) and 269 human instances (i.e., person and rider) that are labeled from 101 real-world foggy images. Furthermore, although there are eight annotated object classes in the Foggy Driving Dataset, we only select the above-mentioned five object classes for detection to ensure consistency between training and testing. 

\item \textbf{RTTS} \cite{li2018benchmarking} is a relatively comprehensive dataset available in natural foggy conditions, which comprises 4322 real-world foggy images with five annotated object classes. Considering hazy/clean image pairs are difficult or even impossible to capture in the real world, Li et al. proposed the RTTS dataset to evaluate the generalization ability of dehazing algorithms in real-world scenarios from a task-driven perspective.
\end{itemize}

\subsection{Implementation Details}
\textbf{Training details}. TogetherNet is trained using the SGD optimizer with a batch size of 16. The initial learning rate $l$ is set to $1 \times 10^{-2}$. We empirically set the total number of epochs to 100 and adopt a Cosine annealing decay strategy to adjust the learning rate $l$. In addition to feeding foggy images to TogetherNet for training the detection task, we also send the original clean images from the VOC-FOG dataset for training the image restoration task. Both the training and testing images are resized to $640 \times 640$. Moreover, we did not add the mosaic data augmentation strategy in the training process as YOLOX defaults, because adopting this approach would increase the difficulty of training our image restoration network, which in turn drops the performance of the object detection task. 

\textbf{Evaluation Settings}. To quantitatively evaluate the performance of the proposed TogetherNet, we adopt the mean Average Precision ($mAP$) as the evaluation metric, which is the most widely used objective evaluation index in object detection tasks. We compare TogetherNet with various state-of-the-art object detection approaches. These object detection methods can be classified into four categories: 1) $``dehaze+detect"$ methods: Here, we employ several dehazing algorithms as a pre-processing step and perform object detection by YOLOXs trained on clean VOC images (the original clean images from VOC-FOG dataset). For pre-processing, we chose four popular dehazing approaches, namely, DCP \cite{he2010single}, AOD-Net \cite{li2017aod}, Semi-dehazing \cite{li2019semi}, and FFA-Net \cite{qin2020ffa} to combine with the YOLOXs detector for forming four combination models called DCP-YOLOXs, AOD-YOLOXs, Semi-YOLOXs, and FFA-YOLOXs, respectively; 2) domain-adaptive-based MS-DAYOLO \cite{hnewa2021multiscale}; 3) multi-task-based DS-Net \cite{huang2020dsnet}; and 4) image adaptive-based IA-YOLO \cite{liu2022image}. Note that all the dehazing algorithms are trained on the entire ITS (Indoor Training Set) \cite{li2018benchmarking} dataset according to the settings in their papers.

\begin{figure*}[!h] \centering
	\includegraphics[width=0.95\linewidth]{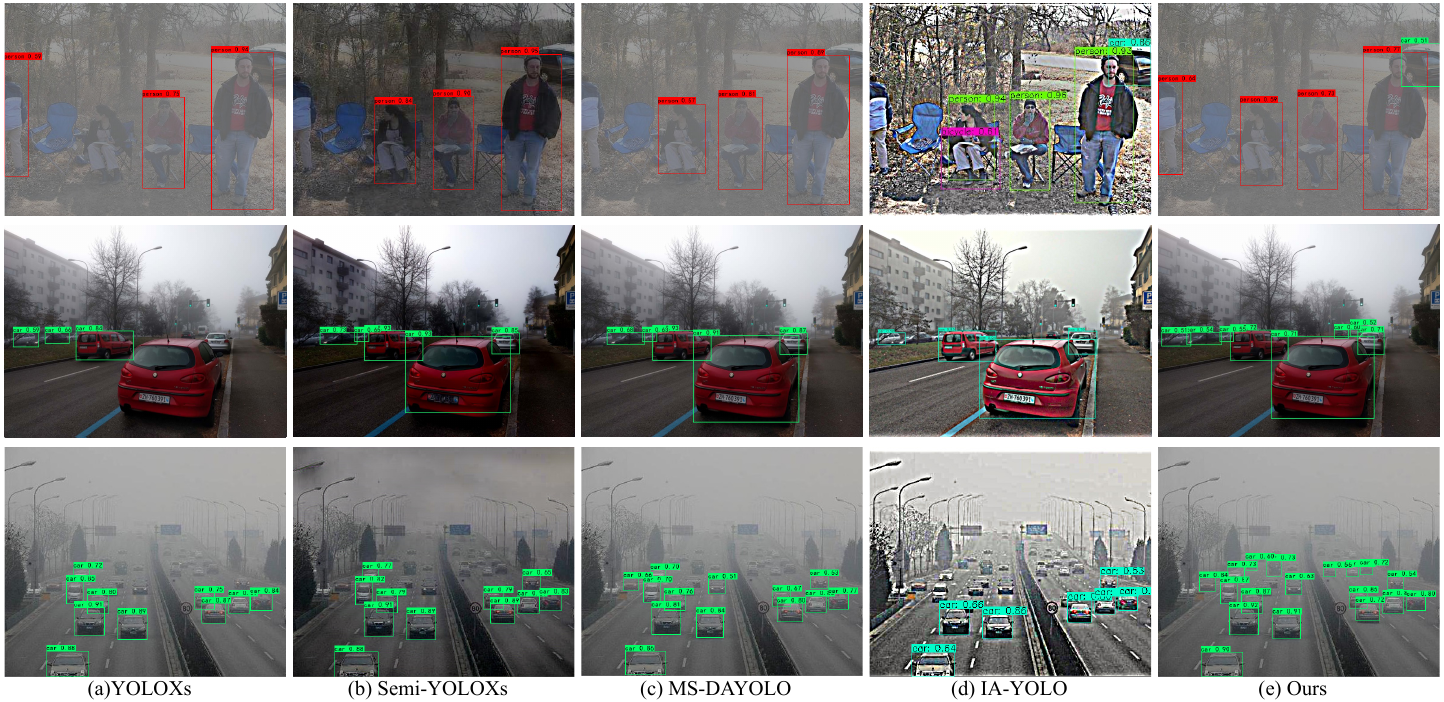}
	\caption{
	Detection results by different methods on both synthetic and real-world foggy datasets. From (a) to (e): the detection results by (a) YOLOXs \cite{ge2021yolox}, (b) Semi-YOLOXs \cite{li2019semi}, (c) MS-DAYOLO \cite{hnewa2021multiscale}, (d) IA-YOLO \cite{liu2022image}, and (e) our TogetherNet, respectively. Clearly, the proposed TogetherNet can discern more objects with higher confidence.}
	\label{fig51}
\end{figure*}

\begin{table}[]
\footnotesize
\caption{Quantitative comparisons (\textit{mAP}) with state-of-the-art detection approaches on the Foggy Driving Dataset.}
  \centering
  \setlength{\tabcolsep}{1.3mm}{
  \begin{tabular}{ccccccc}
    \toprule
    Method  & Person & Bicycle & Car & Motorcycle & Bus & $mAP$ \\
    \midrule
    YOLOXs & 23.26 & 26.55 & 55.04 & 7.14 & 45.71 & 31.54 \\
    YOLOXs* & 26.69 & 23.04 & 56.27 & 2.38 & 41.98 & 30.07 \\
    DCP-YOLOXs* & 22.24 & 10.78 & 56.34 & 7.14 & \textcolor{red}{50.66} & 29.43 \\
    AOD-YOLOXs* & 24.54 & \textcolor{blue}{33.82} & 56.75 & 4.76 & 36.04 & 31.18\\
    Semi-YOLOXs* & 22.39 & 27.73 & 56.47 & 4.76 & 44.93 & 31.26\\
    FFA-YOLOXs* & 19.18 & 18.07 & 50.83 & 2.38 & 42.77 & 26.65\\
    MS-DAYOLO & 21.52 & \textcolor{red}{34.58} & 56.39 & \textcolor{blue}{8.33} & \textcolor{blue}{47.68} & \textcolor{blue}{33.70}\\
    DS-Net & \textcolor{blue}{26.74} & 20.54 & \textcolor{blue}{58.16} & 7.14 & 36.11 & 29.74\\
    IA-YOLO & 16.20 & 11.76 & 41.43 & 4.76 & 17.55 & 18.34\\
    TogetherNet  & \textcolor{red}{30.44} & 28.44 & \textcolor{red}{58.24} & \textcolor{red}{14.29} & 43.23 & \textcolor{red}{34.93}\\
    \bottomrule
  \end{tabular}
  }
  \label{tab3}
\end{table}

\begin{table}[]
\footnotesize
\caption{Quantitative mAP values of the proposed TogetherNet and various state-of-the-art detection approaches on the RTTS dataset. Clearly, our TogetherNet achieves the best performance.}
  \centering
  \setlength{\tabcolsep}{1.3mm}{
  \begin{tabular}{ccccccc}
    \toprule
    Method  & Person & Bicycle & Car & Motorbike & Bus & $mAP$ \\
    \midrule
    YOLOXs & \textcolor{blue}{77.23} & 40.55 & 68.72 & 40.83 & 28.81 & 51.23 \\
    YOLOXs* & 76.07 & 48.47 & 63.88 & \textcolor{blue}{41.03} & 22.76 & 50.44 \\
    DCP-YOLOXs* & 76.81 & \textcolor{blue}{50.03} & 62.84 & 40.62 & 23.73 & 50.81 \\
    AOD-YOLOXs* & 76.49 & 43.32 & 61.03 & 34.54 & 22.16 & 47.51\\
    Semi-YOLOXs* & 75.71 & 46.72 & 62.74 & 40.37 & 24.51 & 50.01\\
    FFA-YOLOXs* & 76.52 & 48.13 & 64.31 & 39.74 & 23.71 & 50.48\\
    MS-DAYOLO & 74.22 & 44.11 & \textcolor{blue}{69.73} & 37.54 & \textcolor{blue}{36.45} & \textcolor{blue}{52.41}\\
    DS-Net & 68.81 & 18.02 & 46.13 & 15.15 & 15.44 & 32.71\\
    IA-YOLO & 67.25 & 35.28 & 41.14 & 20.97 & 13.64 & 35.66\\
    TogetherNet  & \textcolor{red}{82.70} & \textcolor{red}{57.27} & \textcolor{red}{75.32} & \textcolor{red}{55.40} & \textcolor{red}{37.04} & \textcolor{red}{61.55}\\
    \bottomrule
  \end{tabular}
  }
  \label{tab4}
\end{table}

\subsection{Comparison with State-of-the-arts}
\textbf{Comparison on Synthetic Dataset}. The $mAP$ metric of ten detection algorithms on the proposed VOC-FOG-test dataset are reported for quantitative evaluation, as demonstrated in Table \ref{tab2}. To make a fair comparison, we retrain all compared methods (except $``dehaze+detect"$ methods) on the proposed VOC-FOG dataset according to the settings in their papers. For $``dehaze+detect"$ methods, we found that if the baseline YOLOXs is trained on the foggy dataset (VOC-FOG), the final detection results on the testing set will be dropped no matter what dehazing algorithm is employed. This could be an obvious domain shift between the training set (hazy images) and testing set (dehazed images), resulting in a significant decrease in detection accuracy. Therefore, we adopted clean images from the VOC-FOG dataset to train the baseline YOLOXs for these methods. As observed, our TogetherNet outperforms other state-of-the-arts by a large margin in accuracy rate. 

\textbf{Comparison on Real-world Dataset}. We also compare our TogetherNet with several state-of-the-art methods on two real-world foggy datasets, namely, Foggy Driving Dataset and RTTS dataset. Table \ref{tab3} and Table \ref{tab4} exhibit the $mAP$ metric of all compared methods on these two real-world foggy datasets. Different from the results in the VOC-FOG-test dataset, the $``dehaze+detect"$ methods are very limited in improving detection accuracy in real-world degraded scenarios, validating that these processed images do not always guarantee improved object detection performance. clearly, our TogetherNet achieves the highest $mAP$ values again on both datasets, compared to the SOTAs. 

For qualitative comparisons, we exhibit three detection results from the VOC-FOG-test, Foggy Driving, and RTTS datasets in Figure \ref{fig51}. Our TogetherNet is compared with YOLOXs baseline, Semi-YOLOXs, MS-DAYOLO, and IA-YOLO. As observed, TogetherNet can detect more objects with higher confidence, which demonstrates that our approach performs well in both synthetic and real-world foggy datasets. Similar to Semi-YOLOXs, IA-YOLO is also a paradigm of first enhancing the image and then detecting the object, thus they look different from the other approaches.

\begin{table}[]
\footnotesize
\caption{Comparison of TogetherNet with baseline YOLOXs and Syn-YOLOXs ("derain+detect") methods on the RainCityscapes dataset. * denotes that the model is trained with clean images from the RainCityscapes dataset.}
  \centering
  \setlength{\tabcolsep}{1.3mm}{
  \begin{tabular}{ccccccc}
    \toprule
    Method  & Person & Bike & Car & Motorbike & Bus & $mAP$ \\
    \midrule
    YOLOXs & \textcolor{blue}{21.89} & \textcolor{red}{30.37} & 52.36 & 1.26 & 21.15 & \textcolor{blue}{25.41} \\
    YOLOXs* & \textcolor{red}{23.04} & 15.14 & \textcolor{red}{64.47} & 0.03 & 13.61 & 23.26 \\
    Syn-YOLOXs* & 17.11 & 13.29 & 62.22 & \textcolor{blue}{2.03} & \textcolor{blue}{24.55} & 23.84 \\
    TogetherNet  & 19.64 & \textcolor{blue}{19.98} & \textcolor{blue}{64.38} & \textcolor{red}{12.94} & \textcolor{red}{25.28} & \textcolor{red}{28.44}\\
    \bottomrule
  \end{tabular}
  }
  \label{tab888}
\end{table}

\begin{figure}[!h] \centering
	\includegraphics[width=1.0\linewidth]{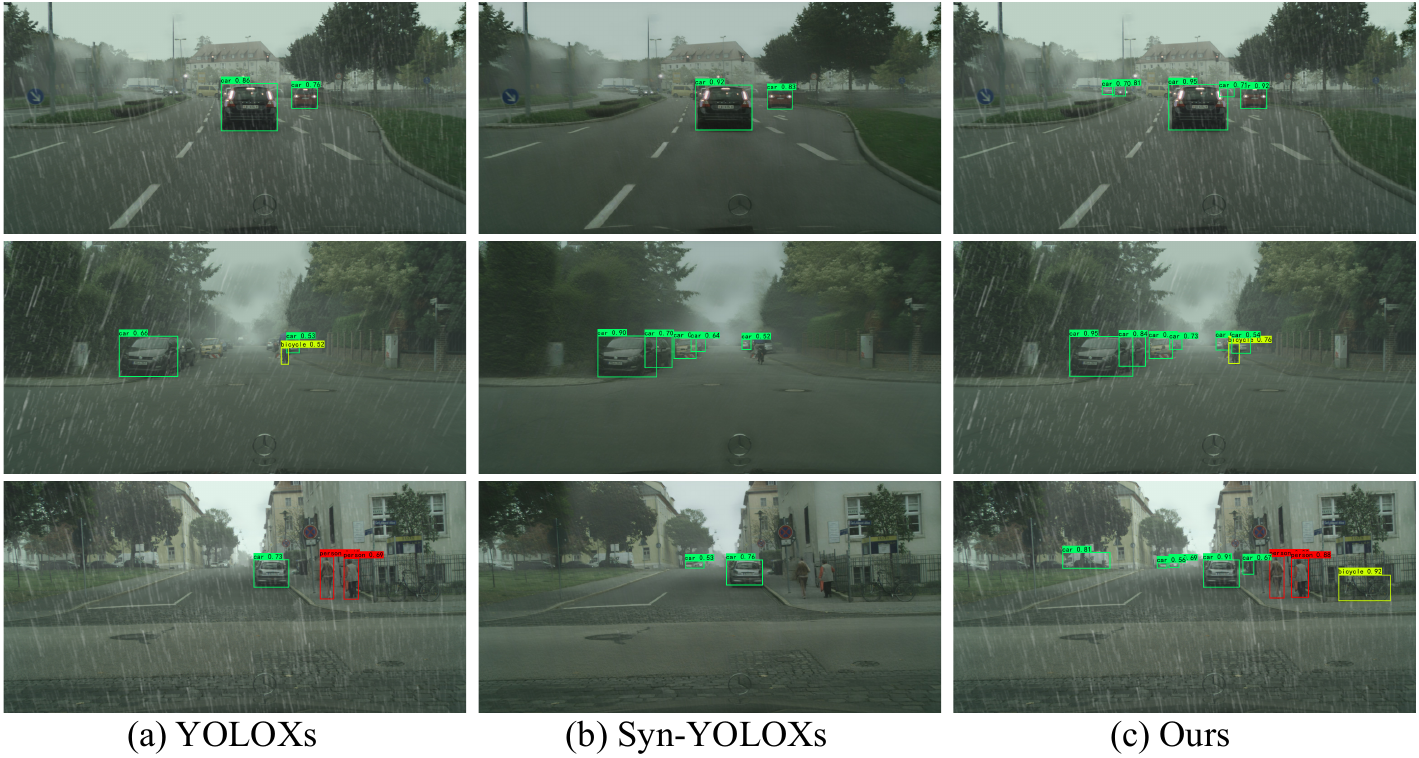}
	\caption{
	Detection results by different methods on the RainCityscapes dataset. From (a) to (c): the detection results by (a) YOLOXs \cite{ge2021yolox}, (b) Syn-YOLOXs \cite{yasarla2020syn2real} ("derain+detect"), and (c) our TogetherNet. }
	\label{fig888}
\end{figure}

\subsection{Experiments on Rainy Images}
To demonstrate that the proposed model can generalize well under other adverse weather conditions, we adopt the RainCityscapes dataset \cite{hu2021single} to evaluate the detection performance of TogetherNet in rainy weather. RainCityscapes dataset contains 10,620 synthetic rainy images (295 images with 36 variants) with eight annotated object classes: car, train, truck, motorbike, bus, bike, rider, and person. In our experiments, we randomly choose 2,500 rainy images (250 images with 10 variants) for training and 450 images (45 images with 10 variants) for testing. As in the previous experimental setup, we only select the aforementioned five object categories for detection.

We compare our TogetherNet with the baseline YOLOXs and a $``derain+detect"$ (Syn2Real \cite{yasarla2020syn2real}) method on the testing set. 
The $mAP$ metric of these 3 detection algorithms is reported in Table \ref{tab888}. As can be seen, our TogetherNet outperforms other approaches by a large margin again in accuracy rate.
Figure \ref{fig888} exhibits three visual examples of the baseline YOLOXs, the $``derain+detect"$ method, and our TogetherNet. As observed, the proposed TogetherNet can discern more objects with higher confidence, which demonstrates that our approach generalizes well under rainy weather conditions.

\begin{table}[]
\centering
\footnotesize
\caption{Ablation study of different training strategies on the RTTS dataset. Clearly, our full model ($V_4$) outperforms other alternatives. IR is the abbreviation of the image restoration module.} 
\begin{threeparttable}
		\centering
		\setlength{\tabcolsep}{1.1mm}{
			\begin{tabular}{ccccccccc}
				\toprule
				\multirow{1}{*}{Variants}&
				\multirow{1}{*}{Base}&
				\multirow{1}{*}{$V_1$}&
				\multirow{1}{*}{$V_2$}&
				\multirow{1}{*}{$V_3$}&
				\multirow{1}{*}{$V_4$}&
				\multirow{1}{*}{$V_5$}&
				\multirow{1}{*}{$V_6$}&
				\multirow{1}{*}{$V_7$} \cr
				\midrule
                IR & w/o & \checkmark & \checkmark & \checkmark & \checkmark & w/o & \checkmark & \checkmark \cr 
                DTFE & w/o & w/o & \checkmark & \checkmark & \checkmark & \checkmark & w/o & \checkmark \cr 
                Focal loss & w/o & w/o & w/o & \checkmark & \checkmark & \checkmark & \checkmark & w/o \cr
                SC Conv & w/o & w/o & w/o & w/o & \checkmark & \checkmark & \checkmark & \checkmark \cr
                \midrule
                $mAP$ & 51.23 & 56.05 & 57.73 & 59.83 & \textbf{61.55} & 54.97 & 56.57 & 56.75\cr
                \bottomrule
			\end{tabular}
		}
\end{threeparttable}
\label{tab5}
\end{table}

\begin{table}[htbp]
	\centering
	\footnotesize
	\caption{Ablation study on the object detection loss $L_{de}$ and image restoration loss $L_{re}$ (loss weight $\lambda_{1}$ and $\lambda_{2}$).}
	\label{tab6}
	\begin{threeparttable}
		\centering
		\setlength{\tabcolsep}{0.9mm}{
			\begin{tabular}{ccccccccc}
				\toprule
				\multirow{1}{*}{$\lambda_{1}$\&$\lambda_{2}$}&
				\multirow{1}{*}{1\&1}&
				\multirow{1}{*}{0.7\&0.3}&
				\multirow{1}{*}{0.5\&0.5}&
				\multirow{1}{*}{0.2\&0.6}&
				\multirow{1}{*}{0.2\&0.8}&
				\multirow{1}{*}{0.2\&1.0}&
				\multirow{1}{*}{0.1\&1.2}\cr
				\midrule
				$mAP$ & 53.79 & 57.44 & 57.73 & 58.30 & \textbf{61.55} & 60.08 & 58.09\cr
				\bottomrule
			\end{tabular}
		}
	\end{threeparttable}
\end{table}

\subsection{Ablation Study} \label{ablation}
\textbf{Effect of different components in TogetherNet}. The proposed network exhibits superior detection performance compared to the state-of-the-art detection methods. To further evaluate the effectiveness of TogetherNet, we conduct extensive ablation studies to analyze the different components, including the image restoration module, Dynamic Transformer Feature Enhancement module (DTFE), Focal loss, and self-calibrated convolutions.

We first construct our base network with the original YOLOXs detector as the baseline of the detection network, and then we train this model with the implementation details mentioned above. Next, we incrementally add different components into the base network as follows:

\begin{figure*}[htbp] \centering
	\includegraphics[width=0.9\linewidth]{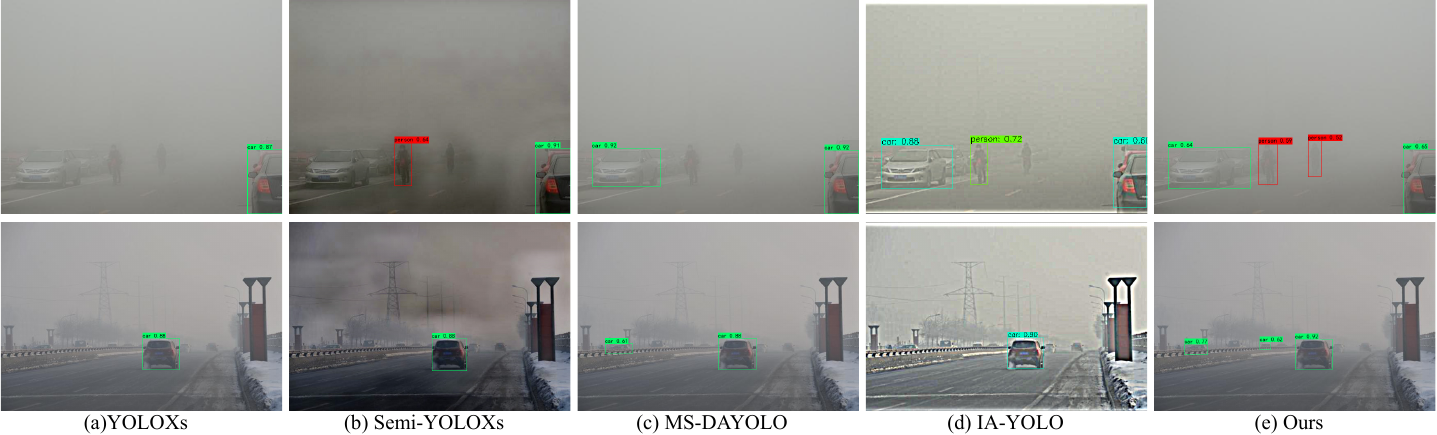}
	\caption{
    Typical failure cases of different detection algorithms. None of the detectors can discern all the objects in such images.}
	\label{fig55}
\end{figure*}

\begin{enumerate}
\item  base model + image restoration module $\rightarrow$ $V_1$, 
\item  $V_1$ + DTFE $\rightarrow$ $V_2$,
\item  $V_2$ + Focal loss $\rightarrow$ $V_3$,
\item  $V_3$ + self-calibrated convolutions $\rightarrow$ $V_4$ (full model),
\item  $V_4$ - image restoration module $\rightarrow$ $V_5$,
\item  $V_4$ - DTFE $\rightarrow$ $V_6$,
\item  $V_4$ - Focal loss $\rightarrow$ $V_7$,
\end{enumerate}
All these variants are retrained in the same way as before and tested on the RTTS dataset. The performances of these models are depicted in Table \ref{tab5}.

As observed, each component in TogetherNet helps in improving object detection performance, especially the proposed image restoration module, which achieves 4.82 $mAP$ gains over our base model. The introduction of the proposed DTFE, Focal loss, and self-calibrated convolutions also greatly improved the performance of the model. In short, if we make full use of the implementation details in this paper, the detection results will outperform other competitive approaches.

\begin{table}[htbp]
\footnotesize
\centering
	\caption{Runtime (in seconds) and \textit{FPS} comparisons of different detection methods tested on an image of 550 $\times$ 400 pixels. * indicates the platform used for image dehazing.}
	\label{Tab1111}
	\begin{threeparttable}
	\centering
 		\setlength{\tabcolsep}{1.9mm}{
			\begin{tabular}{cccc}
				\toprule
				\multirow{1}{*}{Method}&
				\multirow{1}{*}{Platform}&
                \multirow{1}{*}{Run time}&
                \multirow{1}{*}{$FPS$}\cr
				\midrule
				YOLOXs \cite{ge2021yolox}  & PyTorch (GPU)  &  \textcolor{red}{0.018} & \textcolor{red}{55.6}  \cr
				DCP-YOLOXs \cite{he2010single} & Python (CPU)*  & 1.238  & 0.8\cr
				AOD-YOLOXs \cite{li2017aod} & PyTorch (GPU)*  & 0.121 & 8.3\cr
				Semi-YOLOXs \cite{li2019semi} & PyTorch (GPU)* & 1.108 & 0.9\cr
				FFA-YOLOXs \cite{qin2020ffa} & PyTorch (GPU)* & 0.366 & 2.7\cr
				MS-DAYOLO \cite{hnewa2021multiscale} & Caffe (GPU) & 0.037 & 27.0\cr
				DS-Net \cite{huang2020dsnet} & PyTorch (GPU) & 0.035  & 28.6\cr
				IA-YOLO \cite{liu2022image} & Tensorflow (GPU) & 0.039  & 25.6\cr
				TogetherNet  (ours) & PyTorch (GPU)  & \textcolor{blue}{0.031}  & \textcolor{blue}{32.3}\cr
				\bottomrule
			\end{tabular}
		}
	\end{threeparttable}
\end{table}

\textbf{Effect of the weights in loss functions}. To improve the detection performance of TogetherNet in adverse weather conditions, we exploit an effective unified loss function that contains object detection loss $L_{de}$ and image restoration loss $L_{re}$. Accordingly, two loss weights ($\lambda_{1}$ and $\lambda_{2}$) are employed to balance the performance of these two loss functions. For $\lambda_{1}$ and $\lambda_{2}$, extensive experiments are conducted on the RTTS dataset to ensure their optimum values, as exhibited in Table \ref{tab6}. As observed, the image restoration loss is very helpful in improving the detection capacity of the proposed method. Therefore, when setting  $\lambda_{1} = 0.2$ and $\lambda_{2} = 0.8$ in our experiments, the performance of TogetherNet is the best.

\subsection{Efficiency Analysis}
Considering efficiency is essential for a computer vision system, we evaluate the computational performance of various state-of-the-art detection methods and report their average running times and frames per second ($FPS$) metrics in Table \ref{Tab1111}. All the approaches are implemented on a system with an Intel(R) Core(TM) i7-9700 CPU, 16 GB RAM, and an NVIDIA GeForce RTX 3090 GPU. It can be seen that our TogetherNet takes about 0.031$s$ to infer an image of 550 $\times$ 400 pixels on average. The proposed TogetherNet is fast and efficient since it ranks second among the ten detection algorithms.

\subsection{Limitation and Discussion}
Although TogetherNet has achieved encouraging results on both synthetic and real-world foggy datasets, our model is not very robust for the heavily foggy scene. We provide two typical failure cases in Figure \ref{fig55}. It can be observed that the heavy fog degrades the performance of various object detectors. Even humans have difficulty discerning the objects in such challenging images. This limitation might be solved by introducing more effective feature enhancement modules in our network. In near future, we will make efforts to solve this limitation.

\section{Conclusion}
We propose an efficient unified detection paradigm for discerning objects in adverse weather conditions, named TogetherNet. It leverages a joint learning framework to perform image restoration and object detection tasks simultaneously. From a different yet new perspective, TogetherNet casts such detection task as multi-task joint learning, where these two tasks are collaborated and contributed to each other. To better cope with the weather-degradations in this detection task, we develop a Dynamic Transformer Feature Enhancement module (DTFE) to enhance the feature extraction and representation capabilities of our model. In addition, the self-calibrated convolution network is introduced to expand the receptive field of each convolutional layer and enrich the output features, thus reducing the impact of weather-specific information on detection accuracy. Furthermore, we also employ the well-known Focal loss to address the problem of imbalance between positive/negative samples in detection tasks. Experiments on both synthetic and real-world foggy datasets demonstrate that our TogetherNet performs favorably against state-of-the-art detection algorithms.

\vspace{10pt}
\noindent\textbf{Acknowledgements}

 \noindent{This work was supported in part by the National Natural Science Foundation of China (No. 62172218), the Joint Fund of National Natural Science Foundation of China and Civil Aviation Administration of China (No. U2033202), the 14th Five-Year Planning Equipment Pre-Research Program (No. JCKY2020605C003), the Free Exploration of Basic Research Project, Local Science and Technology Development Fund Guided by the Central Government of China (No. 2021Szvup060), the Natural Science Foundation of Guangdong Province (No. 2022A1515010170).}

\bibliographystyle{eg-alpha-doi}
\bibliography{ref}

\end{document}